\documentclass[preprint,12pt]{elsarticle}

\usepackage{amssymb, xcolor, hyperref, bm, subcaption, caption, multirow, amsmath, cleveref, soul, xcolor, pifont, mathabx, booktabs} 

\DeclareMathOperator*{\dist}{dist}
\DeclareMathOperator*{\ssin}{ssin}
\DeclareMathOperator*{\mae}{MAE}
\DeclareMathOperator*{\mse}{MSE}

\newcommand{\mli}[1]{\mathit{#1}} 

\journal{}

\begin{document}

\begin{frontmatter}



\title{A comprehensive study on fidelity metrics for XAI}

\author[1,2]{Miquel Mir\'{o}-Nicolau}
\ead{miquel.miro@uib.es}
\author[1,2]{Antoni Jaume-i-Cap\'{o}}
\ead{antoni.jaume@uib.es}
\author[1,2]{Gabriel Moy\`{a}-Alcover} \corref{cor1}
\ead{gabriel.moya@uib.es}

\cortext[cor1]{Corresponding author}
\address[1]{UGiVIA Research Group, University of the Balearic Islands, Dpt. of Mathematics and Computer Science, 07122 Palma (Spain)}
\address[2]{Laboratory for Artificial Intelligence Applications (LAIA@UIB), University of the Balearic Islands, Dpt. of Mathematics and Computer Science, 07122 Palma (Spain)}



            
\begin{abstract}

The use of eXplainable Artificial Intelligence (XAI) systems has introduced a set of challenges that need resolution. Herein, we focus on how to correctly select an XAI method, an open questions within the field. The inherent difficulty of this task is due to the lack of a ground truth. Several authors have proposed metrics to approximate the fidelity of different XAI methods. These metrics lack verification and have concerning disagreements. In this study, we proposed a novel methodology to verify fidelity metrics, using a well-known transparent model, namely a decision tree. This model allowed us to obtain explanations with perfect fidelity. Our proposal constitutes the first objective benchmark for these metrics, facilitating a comparison of existing proposals, and surpassing existing methods. We applied our benchmark to assess the existing fidelity metrics in two different experiments, each using public datasets comprising 52,000 images. The images from these datasets had a size a 128 by 128 pixels and were synthetic data that simplified the training process. All metric values, indicated a lack of fidelity, with the best one showing a 30 \% deviation from the expected values for perfect explanation. Our experimentation led us to conclude that the current fidelity metrics are not reliable enough to be used in real scenarios. From this finding, we deemed it necessary to development new metrics, to avoid the detected problems, and we recommend the usage of our proposal as a benchmark within the scientific community to address these limitations.

\end{abstract}


\end{frontmatter}


\section{Introduction}

Deep learning models have become ubiquitous solutions and are used across multiple fields, yielding astonishing results. These methods outperform other artificial intelligence (AI) models owing to their high complexity, and ability to learn from large amounts of data. However, this complexity gives rise to a major drawback: the inability to know the reasons behind their results. This challenge is commonly known as the “black-box problem” \cite{barredo_arrieta_explainable_2020}.
 
To address this challenge, eXplainable AI (XAI) has emerged. According to Adadi and Berrada~\cite{adadi2018peeking}, the goal of XAI methods is to “create a suite of techniques that produce more explainable models whilst maintaining high performance levels”. The growing dynamic around XAI has been reflected in several scientific events and the increase in publications as highlighted in several recent reviews about the topic \cite{adadi2018peeking, dosilovic_explainable_2018, murdoch2019interpretable, anjomshoae2019explainable, minh_explainable_2022, barredo_arrieta_explainable_2020, cambria2023survey}. In particular, these methods have been used in sensitive fields such as medical tasks \cite{wang2020covid, wang2017chestx, adarsh2023fair}, where XAI methods are extensively used to gain a deeper understanding of models, improve them, and prevent life-costing mistakes. Multiple methods have emerged to achieve this goal. Murdoch \emph{et al.}~\cite{murdoch2019interpretable} proposed categorizing them into two main categories: model-based and post-hoc. Model-based algorithms refer to AI models that inherently provide insights into the relationships they have learned. The main challenge in model-based explainability lies in developing models that strike a balance between simplicity, making them easily understandable to the audience, and sophistication, enabling them to effectively capture the underlying data. Post-hoc techniques are defined as methods that analyse an externally trained model to provide insights into the learned relationships. These techniques focus on understanding the specific model's behaviour rather than directly interpreting the model's internal mechanisms.

Owing to their simplicity compared with model-based approaches, post-hoc methods have gained widespread adoption, as demonstrated in various studies that reviewed the existing state of the art (\cite{eitel2019testing}, \cite{miro2022evaluating}, \cite{VanderVelden2021}). However, a significant challenge with the post-hoc methods, as highlighted by Adebayo \emph{et al.}~\cite{adebayo2018sanity}, is that different post-hoc methods can produce varying explanations for the same AI model. Krishna \emph{et al.}~\cite{krishna_disagreement_2022} identified and analysed this inconsistency and called it \textit{the disagreement problem}. This problem emphasises the need to identify correct and incorrect explanations to enhance existing techniques. To achieve this, objective evaluation becomes crucial, as relying solely on subjective human evaluation, as stated by Miller \cite{miller2019explanation}, may not yield reliable and consistent results.

Tomsett \emph{et al.}~\cite{tomsett2020sanity} identified fidelity as the main property for detecting whether an XAI algorithm is correct. According to Mohseni \emph{et al.}~\cite{mohseni_multidisciplinary_2021} fidelity is "the correctness of an ad-hoc technique in generating the true explanations (e.g., correctness of a saliency map) for model predictions". The main limitation to calculating it is the inability to have a ground truth of the real explanation. To overcome this limitation, most authors rely on assumptions about the relationship between a correct explanation and the model to measure fidelity. While these proposals may differ in several aspects, all of them involve perturbing the input based on an explanation and analysing the resulting differences in the output of the AI model. Bach \emph{et al.}~\cite{bach2015pixel} proposed perturbing individual pixels from most important to least important and analysing how this perturbation modifies the neural network output, generating a curve that contains the number of pixels perturbed and the output. Samek \emph{et al.}~\cite{Samek2017} used an approach similar to that of Bach \emph{et al.}~\cite{bach2015pixel}, however, instead of perturbing single pixels they perturbed regions of pixels. In addition, they calculated the Area over the Perturbation Curve (AOPC). Rieger \emph{et al.}~\cite{rieger2020irof} modified the proposal of Samek \emph{et al.} defining the regions using a superpixel detection algorithm. Bhat \emph{et al.}~\cite{bhatt2020evaluating} proposed adding perturbations to the original input according to the importance of the explanation, to a completely perturbed image. To obtain the final metric, they proposed using the Pearson correlation coefficient \cite{freedman2007statistics} between the differences in the outputs of the model and the extend of importance that was removed. Alvares-Melis \emph{et al.}~\cite{Alvarez-Melis2018} performed the same calculation as Bhat \emph{et al.}~\cite{bhatt2020evaluating}, with the major difference being that they did not accumulate the perturbation, only perturbing one region at a time. Finally, Yeh \emph{et al.}~\cite{yeh2019fidelity} instead of calculating fidelity proposes calculating the reverse, i.e. infidelity. To do this, they proposed using the expected mean square between the difference in the output when a region is perturbed and the importance of these regions multiplied by the amount of perturbation.  

The existence of numerous fidelity metrics and the absence of a consensus among them pose a significant challenge, which is reminiscent of the disagreement issues identified in XAI methods by Krishna \emph{et al.}~\cite{krishna_disagreement_2022}. In addressing this challenge, several authors have advocated the assessment of metric quality, with Hedström \emph{et al.}~\cite{hedstrom_meta-evaluation_2023} characterising this evaluation as a \textit{meta-evaluation}. To this end, Tomsett \emph{et al.}~\cite{tomsett2020sanity} introduced three sanity checks for fidelity metrics, establishing essential conditions for ensuring their inherent reliability. The application of these checks to the AOPC metric proposed by Samek \emph{et al.}~\cite{Samek2017} and the faithfulness metric proposed by Alvarez-Melis~\cite{Alvarez-Melis2018} revealed that both metrics were deemed "unreliable at measuring saliency map fidelity". Using a methodology similar to that of Tomsett \emph{et al.}~\cite{tomsett2020sanity}, Hedström \emph{et al.}~\cite{hedstrom_meta-evaluation_2023} introduced a set of conditions for accurate measurements, and subsequently converted them into continuous metrics. The application of these metrics to 10 different fidelity measures indicates that Pixel Flipping by Bach \emph{et al.}~\cite{bach2015pixel} performed the best, albeit without achieving perfect results. Both of these assessment approaches can be categorised as axiomatic evaluations, because they establish a set of axioms and assess whether the metrics align with them. However, a noteworthy limitation of these studies lies in the necessity to assume these axioms, especially when a lack of consensus exists between both proposals. Efforts to reconcile and standardise these axioms are imperative for advancing the field and enhancing the reliability of fidelity metrics.

\begin{figure}[!ht]
	\centering
	\subfloat[Flow of a black-box model.]{\includegraphics[width=0.46\textwidth]{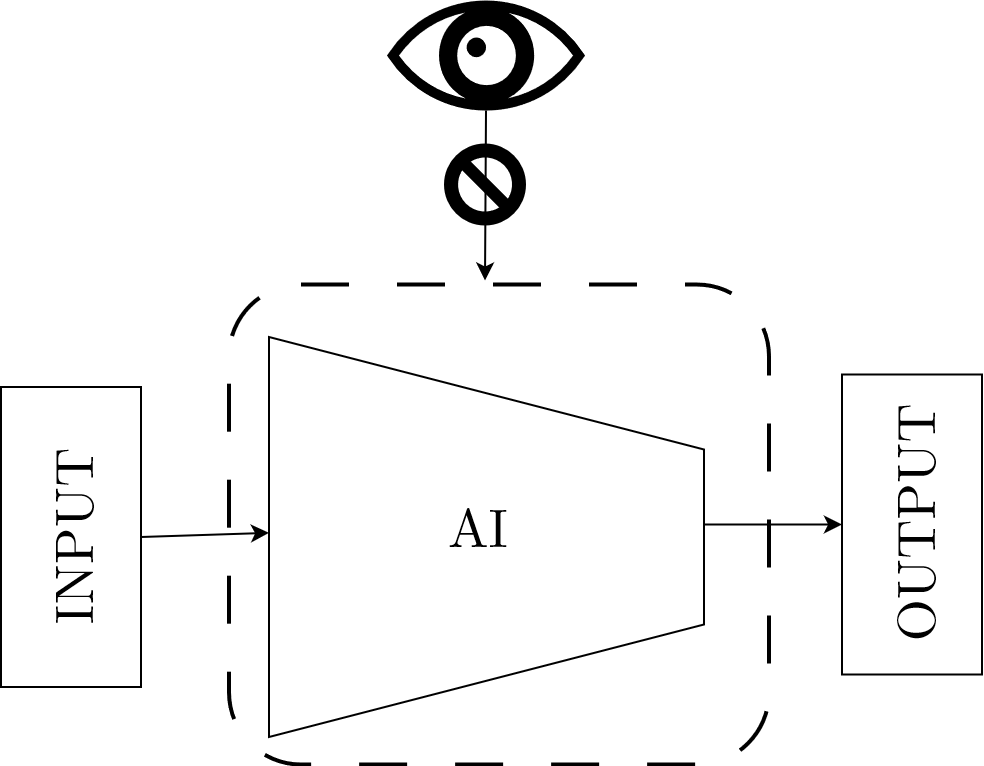}}
	\hfil
	\subfloat[Flow of a black-box model combined with an XAI method.]{\includegraphics[width=0.46\textwidth]{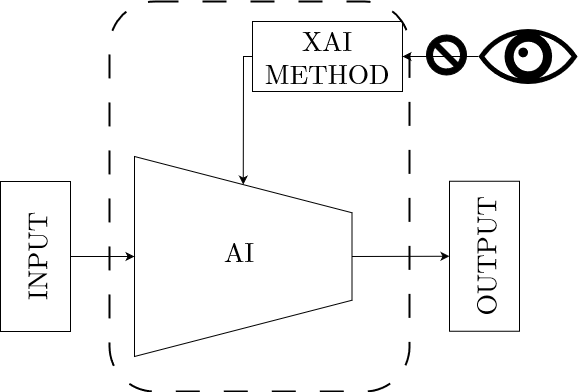}}
	\\
	\subfloat[Flow of a black-box model combined with an XAI method and a fidelity metric.]{\includegraphics[width=0.46\textwidth]{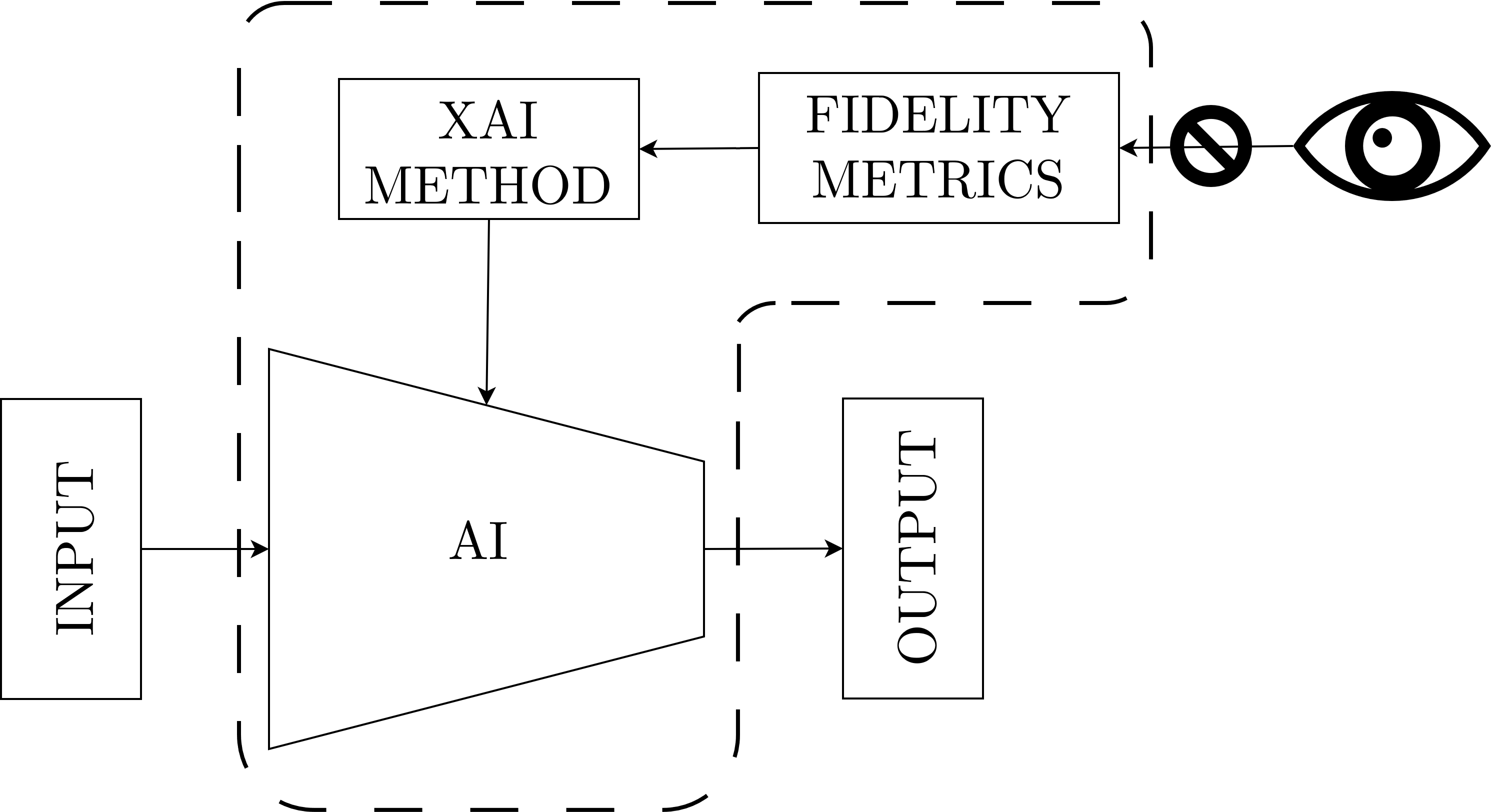}}
    \hfil
 	\subfloat[Flow of a black-box model combined with an XAI method and a verfied fidelity metric.]{\includegraphics[width=0.46\textwidth]{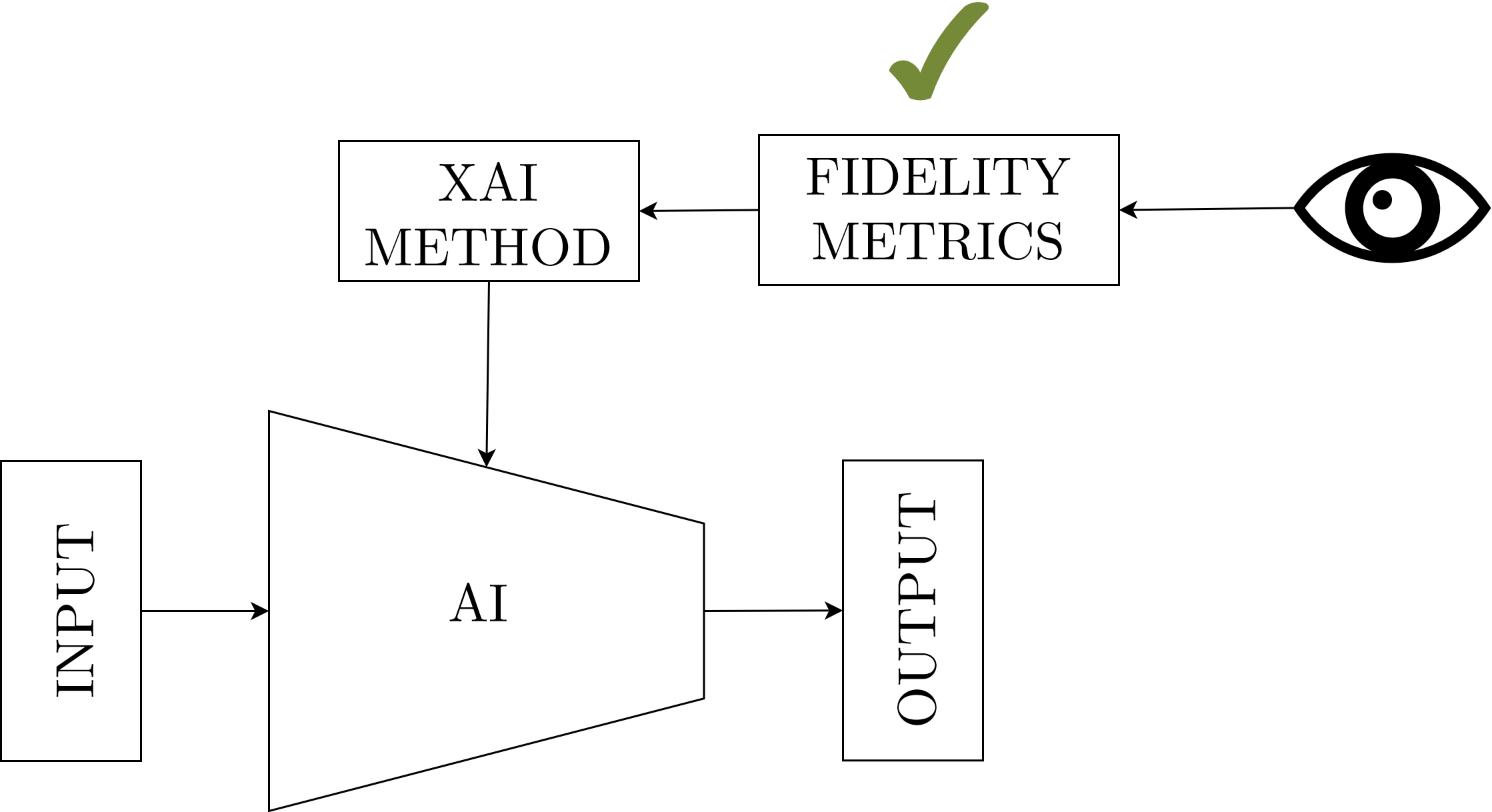}}\\
	\caption{Flows of different configurations: AI model, AI with an XAI method, and AI with an XAI method and a fidelity metric. Inside the dash box, the element that must be trusted is shown.}
	\label{fig:xai_trust}
\end{figure}

From the insights gleaned from these studies, it becomes apparent that the comprehensive XAI methodology does not necessarily eliminate the need for blind trust in black-box models; instead, it introduces its own set of non-transparent elements that demand a similar degree of trust. These components include the AI model itself, the XAI method used, and the fidelity metric employed. Visualised in Figure \ref{fig:xai_trust}, we observe how the inclusion of elements aimed at shedding light on the opaqueness of a pipeline actually only adds complexity to the entire system.

In this study, we aim to develop a novel method to verify fidelity metrics. To accomplish this, we used a transparent model that allowed us to have a ground truth for the explanation: the well-known decision tree \cite{breiman1984classification}. This model allowed us to compare fidelity metrics with the real fidelity of the explanation, surpassing the limitations of the previous approaches and the axiomatic approaches used previously in the literature.

The rest of this paper is organised as follows. In the next section, we identify the objectives of this research. In Section \ref{sec:method} we propose a methodology to measure and analyse the different fidelity metrics. In Section~\ref{sec:experimental_setup}, we specify the experimental environment and describe the fidelity metrics, models, measures, and statistical tests used for experimentation. In Section~\ref{sec:results}, we discuss the results of the two experiments defined in the previous section to analyse the different fidelity metrics, and the theoretical and practical implications of the results. Finally, in Section~\ref{sec:conclusion} we present the conclusions of the study.

\section{Research objectives}

We propose a novel approach to verify the existing fidelity metrics for the XAI methods. These metrics are crucial for a correct XAI system, thereby avoiding the disagreement problem described by Krishna \emph{et al.}~\cite{krishna_disagreement_2022} for XAI methods. However, the reliability of these metrics remains an open question in the current state of the art~\cite{tomsett2020sanity, hedstrom_meta-evaluation_2023}. Therefore, the main goals of the proposed verification metric are as follows: (1) to introduce a novel objective methodology for verifying the reliability of fidelity metrics via the use of a ground truth, which works as the first benchmark for fidelity metrics, and (2) to analyse the existing metric proposals and identify the degree to which they accurately approximate the actual fidelity.

\section{Method} \label{sec:method}

To define our methodology, we firstly formalise the fidelity problem we aimed to discuss. We follow the proposed methodology of Guidotti~\cite{guidotti_evaluating_2021}. 

Let a function $f: \mathcal{X} \rightarrow \mathcal{Y}$ be a model that maps instances $x \in \mathcal{X}$, from the set of possible input data, $\mathcal{X}$, to its respective output $y \in \mathcal{Y}$, where $\mathcal{Y}$ is the set of all ground truths for $\mathcal{X}$. We write $f(x)=y$ to denote the AI result for a particular $x \in \mathcal{X}$. 

These AI models can be classified either as transparent models or black box models. On one hand, transparent models are characterised by knowing the cause behind the decision $f(x)$ for an $x$ input, this cause is known as the explanation, $e_x \in \mathcal{E}$, where $\mathcal{E}$ is the set of all possible explanations. On the other hand, black box models are the model that the explanation is not known. However, the explanation $e_x$ in this kind of models can be approximated by XAI methods, such as $g: \mathcal{X} \times \mathcal{Y} \rightarrow \hat{\mathcal{E}}$, where $\hat{\mathcal{E}}$ is the set of approximations to the original $\mathcal{E}$, and $\hat{e_x}$ is the approximate explanation for an instance $x$. The fidelity of XAI methods with this setup becomes a distance between the real explanation and the approximations, $\dist(\mathcal{E}, \hat{\mathcal{E}})$. The main problem with black-box models is that we did not dispose of $\mathcal{E}$, and for this reason the value $\dist(\mathcal{E}, \hat{\mathcal{E}})$, is calculated via proxy function $\widehat{\dist}(\mathcal{E}, \hat{\mathcal{E}})$. This proxy function is the different fidelity metrics found in the state-of-art \cite{bach2015pixel, Samek2017, rieger2020irof, bhatt2020evaluating, Alvarez-Melis2018, yeh2019fidelity}. 

We proposed to realise the goal of the article, checking whether $\widehat{\dist}(\mathcal{E}, \hat{\mathcal{E}}) \approx \dist(\mathcal{E}, \hat{\mathcal{E}})$. To do so we used transparent models, in which $\mathcal{E} = \hat{\mathcal{E}}$, and for this reason, we know that $\dist(\mathcal{E}, \hat{\mathcal{E}}) = 0$ and that if $\widehat{\dist}(\mathcal{E}, \hat{\mathcal{E}}) \neq 0$ it means that the fidelity metric is incorrect. 

In the following section, we define a set of experiments to check whether the previous requirements are fulfilled in the state-of-art fidelity metrics. 

\section{Experimental setup} \label{sec:experimental_setup}

The experimental setup defined in this section was originally designed to identify the reliability of the fidelity metrics. To do so, we used a transparent model that allowed us to obtain a ground truth for both the  fidelity of the metrics and the explanations.

\subsection{Fidelity metrics}

In the previous section, we analysed the state-of-the-art fidelity metrics. We selected four metrics to further analyse them: Region Perturbation, proposed by Samek \emph{et al.}~\cite{Samek2017}; Faithfulness Correlation, proposed by Bhat \emph{et al.}~\cite{bhatt2020evaluating}; Faithfulness Estimate, proposed by Alvarez-Melis \emph{et al.}~\cite{Alvarez-Melis2018}; and Infidelity, first proposed by Yeh \emph{et al.}~\cite{yeh2019fidelity}.

We discarded the rest of the metrics analysed in the previous section for different reasons: the lack of meaningful differences from the ones selected (Pixel Flipping~\cite{bach2015pixel}, IROF \cite{rieger2020irof} and Selectivity~\cite{montavon2018methods} are similar to Region Perturbation proposed by Samek \emph{et al.}~\cite{Samek2017}) or the nature of the metric (SensitivityN proposed by Ancona \emph{et al.}~\cite{ancona2017towards} is a binary metric, that only indicates whether one result was correct or not). We used implementations from Quantus~\cite{hedstrom2023quantus}.

\subsection{AI model}

We evaluated the four fidelity metrics discussed previously using a transparent model: the regression decision tree. 

This model is a well-known supervised and transparent AI model, based on a tree structure. Its goal is to predict the value of a target variable through binary decision rules inferred from data \cite{breiman1984classification}. These models are extensible and can be used for tabular data; however, in our case, the data we used were images. To use them, we flatten each image and thread it as a flat vector, and each pixel is considered a feature. 

Decision trees are transparent; however, the usual explanations from these models are global ones, with a single explanation for the whole model instead of explaining the decision for one input. Fidelity metrics, in contrast, were designed to analyse local explanations. To obtain a local explanation, we developed a new and simple algorithm. Knowing that the prediction of decision trees is defined by the path from the root node to a leaf node and that this path is selected by analysing a single feature, we proposed to set each of these features as important for prediction. Finally, to quantify this importance, we considered the impurity criterion. As can be seen in Figure \ref{fig:xai_dt}, where a set of examples of explanations are depicted, the result of this process is a sparse explanation, with a very few pixels with any importance. This odd result, compared with usual saliency maps found in the state-of-the-art, was caused by the differences between the convolutional neural networks (the usual model from the saliency maps was extracted) and decision trees:  the former detects local patterns, whereas the latter detects global patterns. Therefore, the saliency maps obtained from decision trees do not highlight local and compact structures, but rather different pixels along the entire image. The algorithm and trained models are available at \url{https://github.com/explainingAI/fidelity_metrics/releases/tag/1.0}.

Owing to the simplicity of decision trees, we have the set of real explanations, $\mathcal{E}$, available. Therefore, the fidelity metric must have perfect results. In other words, in this case, the approximate distance defined by each metric ($\widehat{\dist}$) should be zero because the explanation was perfect. 

\subsection{Datasets}

The experiment presented in this study was based on the use of decision trees, a transparent AI model, which allowed us to obtain explanations with perfect fidelity. This method is not capable of handling complex data such as real images; therefore, we proposed training it using simple synthetic datasets. In particular, we used the AIXI-Shape dataset, proposed by Miró-Nicolau \emph{et al.}~\cite{miro2023novel}, and the $\mli{TXUXIv3}$ dataset proposed by Miró-Nicolau \emph{et al.} in ~\cite{miro2023txuxi}. The original goal of these two datasets was to generate datasets with defined ground truths for the explanations, thus highlighting their simplicity. Both datasets were made public by the authors at \url{https://github.com/miquelmn/aixi-dataset/releases}.

The AIXI-Shape dataset is a collection of 52000 images of 128 by 128 pixels, built by combining a black background and a set of simple geometric shapes (circles, squares, and crosses). Each image varies depending on the position, size, and number of the figures present in it. The label of each image is calculated using equation \ref{eq:ssin},

\begin{equation}
    \ssin(x) = 1/2 \cdot \sin\left(\frac{\pi}{2} |x_{c}|\right) + 1/4 \cdot \sin\left(\frac{\pi}{2} |x_{s}|\right) + 1/6 \cdot \sin\left(\frac{\pi}{2} |x_{cr}|\right),
    \label{eq:ssin}
\end{equation}

where $x$ is an image from the AIXI-Shape dataset, and $|x_{c}|$, $|x_{s}|$, and $|x_{cr}|$ are the number of circles, squares, and crosses present in the image $x$ respectively.

The nature of these images reduces the appearance of out-of-domain (OOD) samples, which is one of the main concerns related to fidelity metrics, owing to the uniform background used. The most common way in which OOD samples are generated, is via the addition of black pixel areas due to the occlusion process, making the background black and mitigating the apparition of known patterns.

The $\mli{TXUXIv3}$ dataset is an extension of the original AIXI-Shape dataset, proposed by the same authors in a different study~\cite{miro2023txuxi}. The authors aimed to generate synthetic images that had the limitations of real datasets, particularly allowing increased OOD generation due to the non-uniform background. This dataset is also a collection of 52,000 images with simple figures, as in the AIXI-Shape dataset, with random locations and sizes. The main difference is the background: instead of a uniform value, the background was randomly selected from 5,640 of the Describable Textures Dataset~\cite{cimpoi14describing}. Similarly to the AIXI-Shape, the label is once again calculated with the $ssin$ function \ref{eq:ssin}.  

\begin{figure}[htb]
	\centering
    	\subfloat{\includegraphics[width=0.24\textwidth]{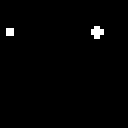}}
    	\hfil
    	\subfloat{\includegraphics[width=0.24\textwidth]{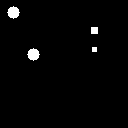}}
        \hfil
    	\subfloat{\includegraphics[width=0.24\textwidth]{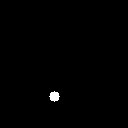}}
    	\hfil
        \subfloat{\includegraphics[width=0.24\textwidth]{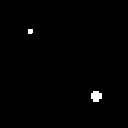}}
     \caption{Sample of images from the AIXI-Shape~\cite{miro2023novel} dataset.}
	\label{fig:aixi_data}
\end{figure}

\begin{figure}[htb]
	\centering
    	\subfloat{\includegraphics[width=0.24\textwidth]{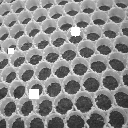}}
    	\hfil
    	\subfloat{\includegraphics[width=0.24\textwidth]{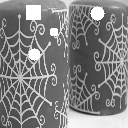}}
        \hfil
    	\subfloat{\includegraphics[width=0.24\textwidth]{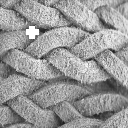}}
    	\hfil
        \subfloat{\includegraphics[width=0.24\textwidth]{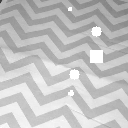}}
     \caption{Sample of images from the $\mli{TXUXIv3}$~\cite{miro2023txuxi} dataset.}
	\label{fig:txuxi_data}
\end{figure}

Examples of these two datasets are shown in Figures \ref{fig:aixi_data} and \ref{fig:txuxi_data}.

\subsection{Experiments}

\begin{figure}[htb]
	\centering
    	\subfloat[Image from the AIXI-Shape~\cite{miro2023novel} dataset.]{\includegraphics[width=0.24\textwidth]{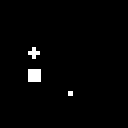}}
    	\hfil
    	\subfloat[Perfect explanation obtained from a decision tree for the previous image.]{\includegraphics[width=0.24\textwidth]{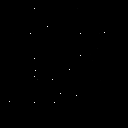}}
        \hfil
    	\subfloat[Image from the AIXI-Shape~\cite{miro2023novel} dataset.]{\includegraphics[width=0.24\textwidth]{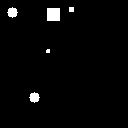}}
    	\hfil
        \subfloat[Perfect explanation obtained from a decision tree for the previous image.]{\includegraphics[width=0.24\textwidth]{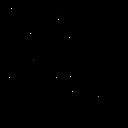}} \\
        \subfloat[Image from the $\mli{TXUXIv3}$~\cite{miro2023txuxi} dataset.]{\includegraphics[width=0.24\textwidth]{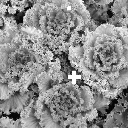}}
    	\hfil
    	\subfloat[Explanation obtained from a decision tree for the previous image.]{\includegraphics[width=0.24\textwidth]{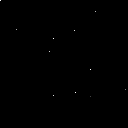}}
        \hfil
    	\subfloat[Image from the $\mli{TXUXIv3}$~\cite{miro2023txuxi} dataset.]{\includegraphics[width=0.24\textwidth]{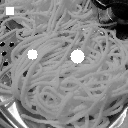}}
    	\hfil
        \subfloat[Explanation obtained from a decision tree for the previous image.]{\includegraphics[width=0.24\textwidth]{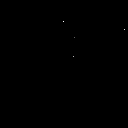}}
     \caption{Examples of images from AIXI-Shape dataset~\cite{miro2023novel}, $\mli{TXUXIv3}$~\cite{miro2023txuxi} dataset and its respective explanations from decision trees.}
	\label{fig:xai_dt}
\end{figure}

We conducted two different experiments to analyse the behaviour of the different fidelity metrics and their reliability. We used the fidelity metrics, AI model and datasets introduced in the previous section. Each experiment aimed to analyse the behaviour of the metrics in a different context, as defined by the data used:

\begin{itemize}
    \item \textbf{Experiment 1}. We trained a decision tree~\cite{breiman1984classification} on the AIXI-Shape dataset, proposed by Miró-Nicolau \emph{et al.}~\cite{miro2023novel}. We used the training and testing divisions from the original dataset: 50,000 images for training and 2,000 for validation. We obtained the local explanations of this transparent model, as explained previously, and calculated the four fidelity metrics on the validation set. Figure \ref{fig:xai_dt} shows a set of images from this dataset and their corresponding explanation. In this experiment, we analysed the behaviour of the fidelity metrics in an environment with fewer OOD samples than usual, which is one of the main concerns of fidelity metrics. We report the mean and standard deviation of the different metrics.    
    \item \textbf{Experiment 2}. Similar to that in the previous experiment, we trained a decision tree~\cite{breiman1984classification}; however, in this case, we used the $\mli{TXUXIv3}$ dataset proposed by Miró-Nicolau \emph{et al.}~\cite{miro2023txuxi}. We used the training and testing division from the original dataset: 50,000 images for train and 2,000 for validation. This dataset, as already discussed, allows for an increased generation of OOD samples due to the presence of a non-uniform background. We analysed the impact of these OOD samples on the fidelity metrics by repeating the same metric calculation as in the previous experiment, using the same method as used previously. Examples of images from this dataset and their corresponding explanations are shown in Figure \ref{fig:xai_dt}
\end{itemize}

In both cases, we used the implementation presented in the \textit{scikit learn} library~\cite{scikit-learn}. We also used the default hyperparameter values from this library. The values can be seen in Table \ref{tab:hyper_params}. We have made the two resulting models publicly available (see \url{https://github.com/explainingAI/fidelity_metrics/releases/tag/1.0}).

\begin{table}[htb]
\centering
\begin{tabular}{lcc}
\toprule
Hyperparameter                                          & Value                 \\ \midrule
Criterion                                               & Gini impurity         \\
Splitter                                                & Best                  \\
Maximum depth                                           & Without maximum       \\
Minimum sample split                                    & $2$                   \\
Minimum samples leaf                                    & $1$                   \\
Minimum weighted fraction leaf                          & $0$                   \\
Maximum features                                        & Number of features    \\
Maximum leaf nodes                                      & Unlimited             \\
Minimum impurity decrease                               & $0$                   \\ \bottomrule
\end{tabular}
\caption{Hyperparameter value for decision tree training in both experiments.}
\label{tab:hyper_params}
\end{table}

The performance of the decision trees in both experiments was not particularly important. The fidelity of the method, which is the main analysis topic of this study, is independent of the performance of the underlying models. A good XAI method must have good fidelity, for both good and bad models. In our case, we assure the fidelity because decision trees are transparent model. However, for the sake of scientific openness, it would be interesting to also have performance measures of the decision trees. We trained these models for a regression task, and used two standard performance measures: Mean Absolute Error (MAE), and Mean Squared Error (MSE) (see equations \ref{eq:mae} and \ref{eq:mse} respectively). Finally, in Table \ref{tab:clf_results} shows the performance measures for the validation set can be seen.

\begin{equation}
    \mae = \frac{\sum^{n}_{i=i} |y_i - \hat{y}_i|}{n},
    \label{eq:mae}
\end{equation}

\begin{equation}
    \mse = \frac{\sum^{n}_{i=i} (y_i - \hat{y}_i)^2}{n},
    \label{eq:mse}
\end{equation}

where $n$ is the size of the dataset, $i$ the index of the image, $y_i$ the prediction of $i$ image, and $\hat{y}_i$ the ground truth of the image $i$.

\begin{table}[htb]
\centering
\begin{tabular}{lcc}
\toprule
Metric                  & Experiment 1 & Experiment 2   \\ \midrule
Mean Absolute Error     & 0.265        & 0.256          \\
Mean Squared Error      & 0.107        & 0.100          \\ \bottomrule
\end{tabular}
\caption{Regression metric performance values for both experiments.}
\label{tab:clf_results}
\end{table}

In neither Experiment 1 nor 2 we compared our approach with any state-of-the-art baseline, because of the novelty of our approach. To the best of our knowledge, this is the first attempt to assess the reliability of fidelity metrics using transparent models with ground truth \cite{hedstrom_meta-evaluation_2023, tomsett2020sanity}.

\section{Results and discussion}\label{sec:results}

In this section, we discuss and analyse the results obtained for the experiments defined in Section~\ref{sec:experimental_setup}.

\subsection{Experiment 1}

Table \ref{tab:first} depicts the results obtained in the first experiment. The table shows the aggregated results for the fidelity metrics with two values: the mean and standard deviation. 

\begin{table}[htb]
\centering
\begin{tabular}{lcc}
\toprule
Metric                                                  & First experiment          \\ \midrule
Faithfulness Correlation~\cite{bhatt2020evaluating}     & $0.7866\; (\pm 0.2963)$   \\
Faithfulness Estimate~\cite{Alvarez-Melis2018}          & $0.7751\; (\pm 0.2888)$   \\
Infidelity~\cite{yeh2019fidelity}                       & $5.9897\; (\pm 23.6442)$  \\
Region Perturbation~\cite{Samek2017}                    & $0.2192\; (\pm 0.1812)$   \\ \bottomrule
\end{tabular}
\caption{Results of the first experiment, obtained using decision trees~\cite{breiman1984classification} for the AIXI-Shape dataset~\cite{miro2023novel}. These results are aggregations of image-wise results: the mean and standard deviation.}
\label{tab:first}
\end{table}

To analyse the results it is crucial to bear in mind that Faithfulness Correlation~\cite{bhatt2020evaluating}, Faithfulness Estimate~\cite{Alvarez-Melis2018}, and Region Perturbation~\cite{Samek2017} are similarity measures, where a value of $1$ represents a perfect result. Conversely, Infidelity~\cite{yeh2019fidelity} is a distance measure, where a value of $0$ indicates perfection, with a possible values range of $[0, +\infty)$   

We can see very different results. Whereas Region Perturbation~\cite{Samek2017} indicated very low fidelity, both Faithfulness Correlation~\cite{bhatt2020evaluating} and Faithfulness Estimate~\cite{Alvarez-Melis2018} indicated much higher fidelity, but still with poor results. The fact that the results of these three fidelity metrics were so different, shows a concerning lack of consensus. Furthermore, these diverging results clearly depict that, in addition to the imperfection of the results, the metrics presents important problems. Finally, we can see that, because of the unbound nature of the metric proposed by Yeh \emph{et al.}~\cite{yeh2019fidelity}, it is more complex to identify whether the explanations are faithful to the real explanation or not. Even so, we found a large dispersion, with a standard deviation of $23.6442$. With values of Infidelity ranging from $0$ (the perfect result) to $481.10$. This large dispersion also indicated that the results depend on the sample used, indicating a clear lack of consistency between samples, in addition to the lack of consensus between metrics.


Although there are big differences between the metrics and some of them were harder to analyse, in our case we know that the explanations have perfect fidelity. Therefore we consider that the results obtained for all four metrics indicated some problems because none of them showed the actual perfect fidelity.

According to the literature~\cite{Gomez2022}, the presence of OOD samples is one of the reasons for erroneous behaviour of occlusion-based approaches, as fidelity metrics, and the fact that the dataset used in this experiment had a fewer OOD samples than usually found in more typical datasets, we expected an even worse result in a real scenario. To confirm this expectation, in the next experiment, we tested the behaviour of these metrics with a dataset with larger appearance of OOD samples.

\subsection{Experiments 2}

Table \ref{tab:second_third_experiment} depicts the results obtained in the second experiment. The table depicts the aggregated metrics for the image-wise results.

\begin{table}[htb]
\centering
\begin{tabular}{lcc}
\toprule
Metric                                                  & Second experiment        \\ \midrule
Faithfulness Correlation.~\cite{bhatt2020evaluating}    & $0.2979\; (\pm 0.3401)$  \\
Faithfulness Estimate~\cite{Alvarez-Melis2018}          & $0.4871\; (\pm 0.3532)$  \\
Infidelity~\cite{yeh2019fidelity}                       & $8.63e7\; (\pm 1.1e10)$  \\
Region Perturbation~\cite{Samek2017}                    & $0.2334\; (\pm 0.1627)$  \\ \bottomrule
\end{tabular}
\caption{Results of the second experiment, obtained using decision trees~\cite{breiman1984classification} for the $\mli{TXUXI}$ dataset~\cite{miro2023txuxi}. These results are aggregations for image-wise results: the mean and standard deviation.}
\label{tab:second_third_experiment}
\end{table}

We observed that all four metrics yielded significantly poorer results than those in the previous experiment, with less fidelity and larger dispersion. These results were obtained in a context in which we knew that the explanation was obtained from a transparent model, and thus hypothetically we expected perfect metrics results for all data. The larger dispersion of all metrics can be seen in, for example, the maximum and minimum values of Infidelity~\cite{yeh2019fidelity}, ranging from $3.09$ to $4.78e10$, which is much worse than those in the previous experiment are.


In the previous experiment, we tested the metrics in a context with fewer OOD samples. However, in this experiment, we used the $\mli{TXUXIv3}$ dataset, proposed by Miró-Nicolau \emph{et al.}~\cite{miro2023txuxi}, which  increases the probability of generating OOD samples because the background is not equal to 0.

We can conclude, based on these results, bearing in mind that the explanation was obtained from transparent models, that the studied fidelity metrics, did not depict the real fidelity of the explanation to the backbone model. The results obtained from these experiments are compatible with those of previous studies that indicated the susceptibility of AI models to OOD samples and the ease of sensitivity approaches, such as fidelity metrics, to generate them~\cite{qiu2021resisting, Gomez2022}. 

\subsection{Theoretical and practical implications}

The proposed methodology allowed objective assessment of the reliability of any fidelity metric. Considering the lack of consensus on how to faithfully calculate the real fidelity of an explanation, a meta evaluation for the metrics, can clearly depict an evaluator correctness can resolve the disagreement problem existing among them. Our proposed methodology can serve as a quality benchmark for future metric developments.

The experimentation in this study used the proposed methodology to compare and analyse the existing fidelity metrics. The results revealed a high sensibility to OOD samples and overall unreliable results, similar to the conclusions obtained in previous axiomatic meta-evaluation proposals~\cite{tomsett2020sanity, hedstrom_meta-evaluation_2023}. All metrics approximate fidelity, obtaining far worse results than the real value.

\section{Conclusion}\label{sec:conclusion}

In this study, we introduced a novel evaluation methodology designed to objectively assess the reliability of fidelity metrics. This evaluation used a transparent model—decision trees—to serve as a quality benchmark for fidelity, due to the inherent availability of a ground truth for the explanation, and consequently for the fidelity.

Using this methodology, we conducted a comprehensive analysis of the current state of fidelity metrics. Specifically, we consolidated all of them into four metrics: Region Perturbation, proposed by Samek \emph{et al.}~\cite{Samek2017}; Faithfulness Correlation, proposed by Bhat \emph{et al.}~\cite{bhatt2020evaluating}; Faithfulness Estimate, proposed by Alvarez-Melis \emph{et al.}~\cite{Alvarez-Melis2018}; and Infidelity, first proposed by Yeh \emph{et al.}~\cite{yeh2019fidelity}. 

Our experimental setup, comprising two distinct experiments, aimed to determine whether existing fidelity metrics accurately reflect the true fidelity of explanations. We hypothesised that accurate metrics would produce impeccable results for transparent decision tree explanations. Contrary to our expectations, none of the metrics consistently delivered perfect outcomes across all samples. Moreover, their performance significantly declined when faced with an increased presence of OOD samples in the second experiment, highlighting their sensitivity to such artefacts.

The susceptibility of fidelity metrics to OOD samples renders them impractical in certain real-world scenarios. In many AI models, one class is designated to include any sample not fitting into other categories. Thus, any perturbation applied to these samples generates new ones that are confidently assigned to this catch-all class. This invariability challenges the explanation of samples within this class through perturbation, revealing an inherent limitation.

In light of these findings, we conclude that the existing state-of-the-art fidelity metrics are ill-suited for accurately calculating explanation fidelity in all practical scenarios, particularly in fields such as  medical related tasks, usually characterised by problems with few classes or even binary, rendering the use of these metrics highly problematic. 

As a future work, and after we have demonstrated that the current fidelity metrics have serious problems, even in a very simple context, our research underscores that it is imperative to develop novel fidelity metrics capable of being correctly used in all scenarios. These new metrics must address the deficiencies inherent in the current approaches and effectively encapsulate the genuine fidelity of explanations. In particular, the lack of reliability of these metrics in the presence of OOD samples, must be fixed. These desiderata can be objectively checked using the proposed methodology. We recommend its use as an initial benchmark to avoid generating more unreliable fidelity metrics.

\section{Declaration of competing interest}

The authors declare that they have no known competing financial interests or personal relationships that could have influenced the work reported in this study.


\section{Funding}
Project PID2019-104829RA-I00 “EXPLainable Artificial INtelligence systems for health and well-beING (EXPLAINING)” funded by \\ MCIN/AEI/10.13039/501100011033. Miquel Miró-Nicolau benefited from the fellowship FPI\_035\_2020 from Govern de les Illes Balears.

\bibliographystyle{apalike} 
\bibliography{references}





\end{document}